\documentclass[runningheads]{llncs}
\usepackage{graphicx}
\usepackage{amsmath}
\usepackage{amssymb}
\usepackage[ruled,vlined]{algorithm2e}
\usepackage{graphicx}
\usepackage[utf8]{inputenc}
\usepackage{float}
\usepackage{mathabx}
\usepackage[T1]{fontenc}
\begin{document}
%
\title{MAGIC: Multi-scale Heterogeneity Analysis and Clustering for Brain Diseases}
%
%
 \author{Junhao Wen\inst{1}\orcidID{0000-0003-2077-3070} \and
 Erdem Varol\inst{2}\orcidID{0000-0002-6128-8939} \and
 Ganesh Chand\inst{1,3} \and
 Aristeidis Sotiras \inst{4}\orcidID{0000-0003-0795-8820} \and
 Christos Davatzikos\inst{1}\orcidID{0000-0002-1025-8561}}
 
\index{Wen, Junhao}
\index{Varol, Erdem}
\index{Chand, Ganesh}
\index{Sotiras, Aristeidis}
\index{Davatzikos, Christos}

\authorrunning{J. Wen et al.}
\titlerunning{ }
%
\institute{Center for Biomedical Image Computing and Analytics, Perelman School of Medicine, University of Pennsylvania, Philadelphia, USA\\
\email{junhao.wen89@gmail.com}\\›
\url{https://www.med.upenn.edu/cbica/} \and
Department of Statistics, Center for Theoretical Neuroscience, Zuckerman Institute, Columbia University, New York, USA \and
Department of Radiology, School of Medicine, Washington University in St. Louis, St. Louis, USA \and
Department of Radiology and Institute for Informatics, Washington University School of Medicine, St. Louis, USA}

\maketitle              
\begin{abstract}
There is a growing amount of clinical, anatomical and functional evidence for the heterogeneous presentation of neuropsychiatric and neurodegenerative diseases such as schizophrenia and Alzheimer's Disease (AD). Elucidating distinct subtypes of diseases allows a better understanding of neuropathogenesis and enables the possibility of developing targeted treatment programs. Recent semi-supervised clustering techniques have provided a data-driven way to understand disease heterogeneity. However, existing methods do not take into account that subtypes of the disease might present themselves at different spatial scales across the brain. Here, we introduce a novel method, MAGIC, to uncover disease heterogeneity by leveraging multi-scale clustering. We first extract multi-scale patterns of structural covariance (PSCs) followed by a semi-supervised clustering with double cyclic block-wise optimization across different scales of PSCs. We validate MAGIC using simulated heterogeneous neuroanatomical data and demonstrate its clinical potential by exploring the heterogeneity of AD using T1 MRI scans of 228 cognitively normal (CN) and 191 patients. Our results indicate two main subtypes of AD with distinct atrophy patterns that consist of both fine-scale atrophy in the hippocampus as well as large-scale atrophy in cortical regions. The evidence for the heterogeneity is further corroborated by the clinical evaluation of two subtypes, which indicates that there is a subpopulation of AD patients that tend to be younger and decline faster in cognitive performance relative to the other subpopulation, which tends to be older and maintains a relatively steady decline in cognitive abilities.
\end{abstract}

\section{Introduction}
Imaging patterns of various brain diseases, such as schizophrenia (SCZ) \cite{SCZ_biomarker,SCZ_biomarker2,SCZ_biomarker3} and Alzheimer's Disease (AD) \cite{AD_biomarker,AD_biomarker2,AD_biomarker3} are often investigated via group comparisons involving voxel-based or vertex-based statistical analyses. However, such approaches typically assume that a unique pathological pattern exists in the disease group and are agnostic to the potential heterogeneity of neuropathogenesis due to unobserved endophenotypes. Ignoring heterogeneity may lead to underpowered statistical conclusions due to the violation of the unimodality assumption of effect loci in the group comparisons. 
\par
Several previous studies have made efforts to reveal the heterogeneous clinical biomarkers by leveraging machine learning (ML) and neuroimaging techniques. These studies can be generally divided into two classes based on whether the data clustering is unsupervised or semi-supervised. Unsupervised clustering \cite{AD_subtypes,AD_subtype2,AD_subtype3} aims to directly cluster the patients with regard to their demographic information, clinical presentation or imaging biomarkers. However, unsupervised clustering techniques rely on similarity or dissimilarity measures across the patient group only, which can potentially be confounded by covariate effects such as age, sex and other sources of variation that are not related to the disease effect. These confounds may overpower and mask the true heterogeneous pathological effects caused by the disease. Moreover, the optimal number of clusters (\emph{c}) is often set apriori, instead of being determined by cross-validation (CV). Alternatively, several recent techniques have been proposed to utilize semi-supervised clustering to distinguish heterogeneous disease effects. In \cite{HYDRA}, the authors propose a method, termed HYDRA, to discriminate between controls and patients using a convex-polytope classifier while simultaneously clustering the patient subgroups. The covariate effects, such as age and sex, are regressed out and the optimal number of clusters is decided via a CV procedure. Moreover, the authors demonstrate HYDRA’s superiority over other unsupervised methods, such as K-means. However, HYDRA performs clustering inference using an input set of features that reflect a single spatial scale of anatomy captured by an apriori determined set of regions of interest (ROI) and may not be able to capture heterogeneous patterns that span a wider spectrum of spatial scales. This may lead to an inaccurate exposition of the heterogeneous disease patterns presented in the clinical study.
\par
To address this limitation, we propose a novel method, Multi-scAle heteroGeneity analysIs and Clustering (MAGIC)\footnote[1]{https://github.com/anbai106/MAGIC}, for parsing multi-scale disease heterogeneity. MAGIC first extracts multi-scale, from macro to micro, patterns of structural covariance (PSCs), analogously with atlas-based ROIs, via orthogonal projective non-negative matrix factorization (OPNMF) \cite{OPNMF}. Then a semi-supervised clustering procedure through a double cyclic block-wise optimization \cite{Coordinate_descent} is leveraged to yield robust clusters. Furthermore, MAGIC allows us to obtain a data-driven parcellation that can explain the heterogeneity the most, thus can also be seen as a heterogeneity aware segmentation technique.

To demonstrate our claims, we first validate MAGIC on simulated effect data with the known number of clusters and multi-scale atrophy patterns. Here we show that MAGIC recovers both the underlying imaging patterns and the correct number of clusters. We then apply MAGIC to ADNI data to disentangle the heterogeneity of AD which reveals two distinct subtypes, where one subtype presents macro-scale cortical atrophy while the latter subtype exhibits focused hippocampus atrophy.

\section{Method}
All T1-weighted (T1w) MR images underwent the following image preprocessing procedure: Brain tissue segmentation was performed using a multi-atlas segmentation technique \cite{MUSE} and was then transformed to produce tissue density maps. Gray matter (GM) tissue density maps were smoothed and harmonized by estimating age and gender effects in CN using a voxel-wise linear model.
\par
\begin{figure}[!tb]
\centering
\includegraphics[width=\textwidth]{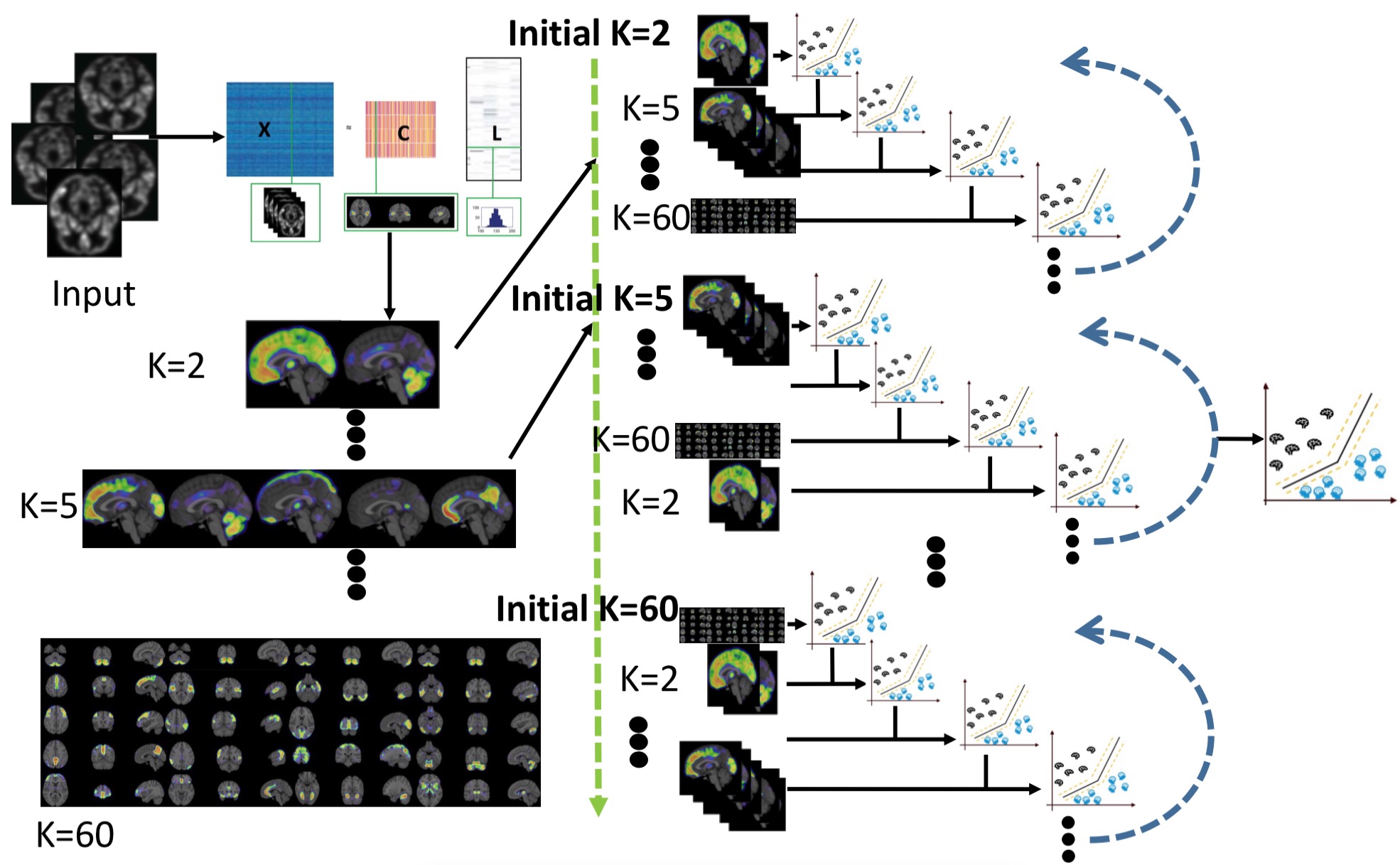}
\caption{The schematic diagram of MAGIC.} \label{fig1}
\end{figure}

The schematic diagram of MAGIC is shown in Fig.~\ref{fig1}. Let the input data (i.e., GM density maps) consist of \emph{N} subjects, with \emph{D} features for each subject. All participants are labeled as 1 for AD and -1 for CN. The input data are denoted as: \(\boldsymbol{X} = (\boldsymbol{x}_i, y_i)_{i=1}^N \quad (\boldsymbol{X} \in \mathbb{R}^{D \times N} \quad \textrm{and} \quad y \in \{-1, 1\}\)). 

\subsection{Representation Learning for Multi-scale PSCs Extraction}
In MAGIC, orthogonal projection NMF \cite{OPNMF} is used for multi-scale PSCs extraction. Low dimensional features can be extracted from coarse to refined scales with different predefined number of PSCs (\emph{K}). The general form of the NMF can be defined as: 
\begin{equation}
\begin{split}
\|\boldsymbol{X} - \boldsymbol{C}\boldsymbol{L}\|_F^2~~~ \textrm{subject to} \quad \boldsymbol{C} \geq 0,~~\boldsymbol{L}\geq 0,~~ \boldsymbol{C}\boldsymbol{C}^T = \boldsymbol{I}
\end{split}
\end{equation}
where matrix $\boldsymbol{C} = [\boldsymbol{c}_1, ..., \boldsymbol{c}_K]$ contains the \emph{K} estimated PSCs. $\boldsymbol{c}_i \in \mathbb{R}^D $ and is assumed to be a unit vector $\|\boldsymbol{c}_i\|^2=1 $. $\boldsymbol{I}$ represents the identity matrix. We refer $\boldsymbol{C} \in \mathbb{R}^{D\times K}$ as component matrix and $\boldsymbol{L} \in \mathbb{R}^{K\times N}$ as loading coefficient matrix. Both $\boldsymbol{C}$ and $\boldsymbol{L}$ are non-negative and indispensable to approximate the original data $\boldsymbol{X}$. Apart from being non-negative, another constraint we explicitly impose is that $\boldsymbol{L}$ is estimated as the orthogonal projection of the input $\boldsymbol{X}$ to the components $\boldsymbol{C}$ $(\boldsymbol{L} = \boldsymbol{C}^T\boldsymbol{X}$).
\par
The component matrix is a sparse part-based representation and conveys the information regarding the spatial properties of the variability effect. On the other hand, the loading coefficient matrix is a low level feature representation which quantifies the strength of those spatial properties in each data sample. In the current work, we take \emph{K} from 2 to 60, resulting in 59 sets of single-scale PSCs and 1829 PSCs in total.  

\subsection{Clustering via Max-margin Multiple SVM Classifiers}
MAGIC constructs the convex-polytope classifier in the same way as HYDRA does \cite{HYDRA}. For each clustering subproblem, MAGIC takes a specific scale of PSCs as input features (\(\boldsymbol{L}^T, \boldsymbol{L} \in \mathbb{R}^{K \times N}\)) and corresponding \(y \in \{-1, 1\}\) as labels. 
\par 
In a nutshell, the polytope in the search space is made up by all support vector machine (SVM) hyperplanes: each hyperplane contributes to one face of the polytope. Without loss of generality, let us confine CN to be in the interior of the polytope. MAGIC aims to correctly classify all CN and at least one SVM correctly classify each patient. The objective of maximizing the polytope's margin can be summarized as:
\begin{equation}
\begin{split}
& \max_{\{\boldsymbol{w}_j, b_j\}_{j=1}^c} \frac{1}{c}\sum_{j=1}^{c}\frac{2}{\|\boldsymbol{w}_j\|_2}\\
&\textrm{subject to} \quad \boldsymbol{w}_{j}^T\boldsymbol{L}^T_{i} + b_j \leq -1, \textrm{if} \quad y_j = -1;~~\boldsymbol{w}_{j}^T\boldsymbol{L}^T_{i} + b_j \geq 1, \textrm{if} \quad y_j = 1
\end{split}
\end{equation}
where \({\boldsymbol{w}_j}\) and \({\boldsymbol{b}_j}\) are the weight and bias, respectively and are sufficient statistics to define the faces of the convex polytope. In general, this optimization routine is non-convex and is solved by iterating on solving for cluster memberships and solving for polytope faces' parameters~\cite{HYDRA}.
\subsection{Double Cyclic Block-wise Optimization}
MAGIC adopts a double cyclic block-wise optimization procedure in order to combine the knowledge from different scales of PSCs. The block-wise optimization solves the clustering problem in the form of
\begin{equation}
\begin{split}
& \max_{{(\boldsymbol{w}_1, {b}_1)}, ..., {(\boldsymbol{w}_a, {b}_a)}}     \digamma({(\boldsymbol{w}_1, {b}_1)}, ..., {(\boldsymbol{w}_a, {b}_a)}) \\
\end{split}
\end{equation}
where \(a\in \mathbb{R}\) is the number of iterations/blocks that the optimization takes until the predefined stopping criterion achieves. \({(\boldsymbol{w}_a, {b}_a)}\) is the weight and bias term derived by the a-th set of PSCs. 
\par
The cyclic block-wise optimization (i.e., blue dotted arrow in Fig.~\ref{fig1}) aims to minimize each specific set of single-scale PSCs (\(\digamma({\boldsymbol{w}_a, {b}_a})\)), while fixing the remaining blocks. The model is first initialized from a specific set of \(K\) PSCs. Then the model is transferred to the next block for fine-tuning the polytope. This updating rule was performed in a cyclic order across different \emph{K} until consistent clustering results were obtained across scales. This cyclic procedure can be summarized in the form of  
\begin{equation}
\begin{split}
& {\boldsymbol{S}_1}   \triangleq \digamma({\boldsymbol{w}_1, {b}_1}) \\
&...\\
& {\boldsymbol{S}_a}   \triangleq \digamma(\digamma({\boldsymbol{w}_1, {b}_1})...\digamma({\boldsymbol{w}_a, {b}_a}))                                 \\
\end{split}
\end{equation}
where \(\boldsymbol{S}_a\) is the search space for the convex-polytope. The second loop (i.e., green dotted arrow in Fig.~\ref{fig1}) is to initialize the polytope with different \(K\), in order to achieve a consensus clustering solution. 

\section{Experiments}
For synthetic data, 364 CN from a healthy control dataset were included and randomly split into two half-split sets. The first split was defined as CN and the second as a pseudo patient group. The pseudo patient group was further divided into two half-split sets for neuroanatomical heterogeneity simulation: Global cortical and subcortical atrophy were introduced to the first and second splits, respectively. Atrophy simulation with 10\% voxel-wise intensity reduction was imposed to the predefined regions for each PT splits. For real data, we applied MAGIC to ADNI 1 data with 228 CN and 191 AD. 
\par
The number of clusters \emph{c} was decided by CV. Stratified repeated holdout splits \cite{sklearn,CV_Varoquaux} with 100 repetitions were performed and adjusted rand index (ARI)\cite{ari_hubert} was used to quantify the clustering stability. For simulated data, we compare clustering performance and evaluate classification task performance by using balanced accuracy (BA). Moreover, each synthetic clustering experiment was repeated 50 times.
\par
Statistical mapping was performed between the subtypes and CN. A two-sample t-test was performed for all 1829 PSCs with the significance level as 0.05. Benjamini-Hochberg (BH) adjustment \cite{BH} was used for the correction of multiple comparisons. Furthermore, the effect size (ES) for Cohen's \emph{d} \cite{cohend} was computed for those PSCs that survived the correction.

\section{Results}
\subsection{Synthetic Experiments}
\subsubsection{Clustering Stability via Cross-validation}
\begin{figure}[!tb]
\centering
\includegraphics[width=\textwidth]{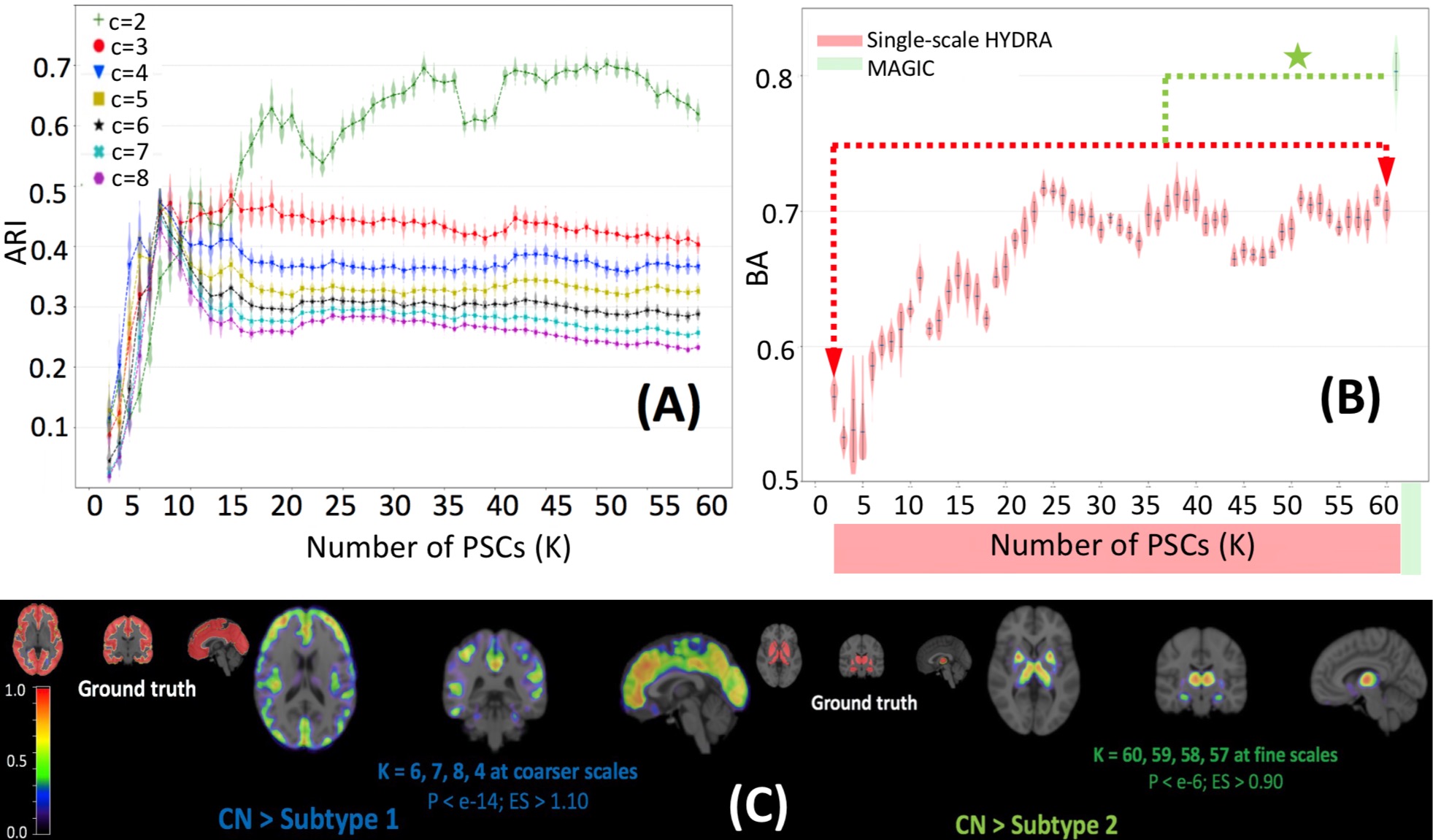}
\caption{Synthetic results. (A): Clustering stability across number of clusters (\emph{c} = 2 to 8) and number of PSCs (\emph{K})= 2 to 60). (B): Clustering performance comparison for different approaches: Single-scale HYDRA (in red) and MAGIC (in green). The green star denotes for statistically significant difference. (C): Statistical mapping between subtypes and CN with MAGIC results.} \label{fig2}
\end{figure}

Single-scale HYDRA was first applied to the synthetic data for choosing optimal \emph{c}. Fig.~\ref{fig2}A shows the clustering stability for different \emph{c} (\emph{c} = 2 to 8). In general, unstable phase (\emph{K} = 2 to 16) gives mixed ARIs across different \emph{c}. In stable plateau phase (\emph{K} = 25 to 60), \emph{c} equals 2 obtains consistent higher ARIs than other \emph{c}. The stable plateau phase scales (\emph{K} = 25 to 60 with step size = 5) were then used for MAGIC. 

\par 
\subsubsection{Clustering Performance Comparison between Approaches}
The comparison of clustering performance between approaches was shown in Fig.~\ref{fig2}B. Overall, single-scale HYDRA obtained inferior performance and the highest mean BA was achieved at \emph{K} = 38 (\(0.71\pm0.009\)). MAGIC used multi-scale PSCs (\emph{K} = 25 to 60 with step size = 5) and achieved statistically higher BA compared to single-scale HYDRA (i.e., mean BA = \(0.81\pm0.014\), p-value \(\lll\) 0.05). Of note, fitting all 1829 PSCs to HYDRA does not give comparable results. 

\par 
\subsubsection{Neuroanatomical Heterogeneity between Subtypes and CN}
The neuroanatomical spatial patterns based on MAGIC clustering results are displayed in Fig.~\ref{fig2}C. We presented only the PSCs with highest ES, and the corresponding P-value and ES. 
\par 
For subtype 1, diffuse cortical atrophy was observable: 1446 out of the 1829 PSCs showed significant difference. Among those 1446 PSCs, the PSC covering almost the whole cortical regions showed highest ES (1.10). Note that this PSC was simultaneously extracted from multiple coarser scales (e.g.,\emph{K} = 6, 7, 8, 4). For subtype 2, focal subcortical atrophy was found. 22 out of the 1829 PSCs were significantly different. Similarly, the 22 PSCs were the same component from different \emph{K}, which encompassed the subcortical structures (i.e., hippocampus, thalamus, putamen and caudate).     
\subsection{Alzheimer's Disease Dataset Experiments}
\subsubsection{Clustering Stability via Cross-validation} 
\begin{figure}[!tb]
\centering
\includegraphics[width=\textwidth]{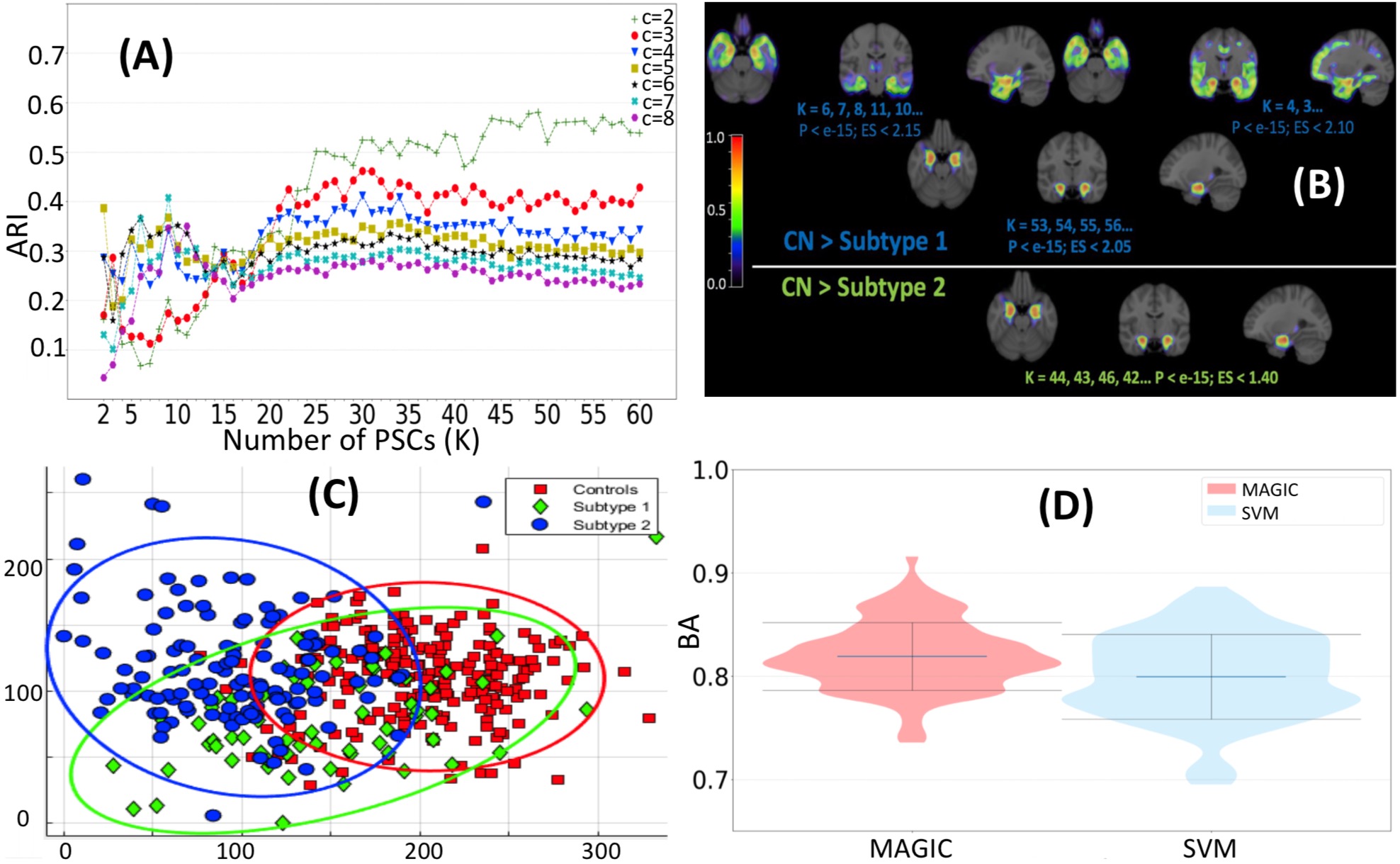}
\caption{ADNI data results. (A): Clustering stability across number of clusters (\emph{c} = 2 to 8) and number of PSCs (\emph{K})= 2 to 60). (B): Statistical mapping between subtypes and CN with MAGIC. (C): 2D multidimensional scaling visualization of the subgroups (in blue and green) relative to controls (in red) using the top two features that is used in MAGIC clustering. (D): Classification results for MAGIC polytope and linear SVMs, respectively.} \label{fig3}
\end{figure}

Fig.~\ref{fig3}A shows the clustering stability for different \emph{c} (\emph{c} = 2 to 8) for single-scale HYDRA. Unstable phase gave mixed ARIs across different \emph{c} (\emph{K} = 2 to 20). In stable plateau phase (\emph{K} = 25 to 60), \emph{c} equals 2 obtained consistent higher ARI than other \emph{c}. Thus we chose \emph{c} = 2 to be the optimal number of clusters. The stable plateau phase scales (\emph{K} = 25 to 60 with step size = 5) were subsequently used for MAGIC. 

\subsubsection{Demographic and Clinical Characteristics of clustering subtypes}
Table~\ref{tab1} displayed the demographic and clinical characteristics for ADNI participants and the corresponding subtypes based on MAGIC. Age and FDG are significantly different between two subtypes at baseline. MMSE and ADAS become significantly different between subtypes changing from baseline to 12 months.  
\begin{table}[!tb]
\caption{Demographic and Clinical Characteristics of clustering subtypes. AD patients (left) and the estimated subtypes of AD (right). APOE4 denotes subjects with at least one APOE allele present. M12 and bl represent time point at 12 months and baseline, respectively. * denotes statistical significance.}\label{tab1}
\begin{tabular}{|l|l|l|l|l|l|l|}
\hline
Characteristics &  CN(n=228) & AD(n=191) & Pvalue & Sub1(n=134) & Sub2(n=57) & Pvalue \\
\hline
Age (years) & 75.87$\pm$5.03 & 75.27$\pm$7.46  & 0.32 & 73.96$\pm$7.46 & 78.34$\pm$6.56 & 6.00e-4* \\
Sex (female) & 110 (48.25) & 91 (47.64) & 0.98 & 62 (46.27) & 29 (50.88) & 0.67 \\
APOE4 & 61 (26.75) & 100 (52.36) & 5.13e-17* & 90 (67.16) & 37 (64.91) & 0.15\\
FDG & 6.41$\pm$0.61 & 5.39$\pm$0.67  & 1.19e-22* & 5.21$\pm$0.63 & 5.74$\pm$0.62 & 2.08e-04*\\
MMSE bl & 29.11$\pm$1.00 & 23.31$\pm$2.04  & 6.03e-137* & 23.31$\pm$2.02 & 23.29$\pm$2.10 & 0.96\\
MMSE M12 & 29.13$\pm$1.17 & 22.69$\pm$4.08  & 3.73e-53* & 20.40$\pm$4.44 & 22.69$\pm$4.08 & 2.58e-03*\\
ADAS11 bl & 6.21$\pm$2.92 & 18.67$\pm$6.25  & 1.07e-92* & 18.86$\pm$6.01 & 18.29$\pm$6.72 & 0.55\\
ADAS11 M12 & 5.52$\pm$2.86 & 22.66$\pm$9.38  & 4.92e-81* & 24.01$\pm$8.92 & 19.51$\pm$9.76 & 5.06e-03*\\
ADAS13 bl & 9.50$\pm$4.19 & 28.97$\pm$7.57  & 4.64e-118* & 28.23$\pm$8.20 & 29.34$\pm$7.23 & 0.35\\
ADAS13 M12 & 8.79$\pm$4.58 & 33.47$\pm$10.89  & 1.24e-97* & 35.11$\pm$10.25 & 29.78$\pm$11.48 & 4.44e-03*\\

\hline
\end{tabular}
\end{table}
\subsubsection{Neuroanatomical Heterogeneity between Subtypes and CN}
Fig.~\ref{fig3}B shows the neuroanatomical spatial patterns for MAGIC results. Two subtypes showed distinct atrophy patterns. For subtype 1, diffuse atrophy pattern was established on the whole brain: 1560 out of the 1829 PSCs showed significant difference. Those PSCs with the highest ES included hippocampus, temporal and frontal lobe. For subtype 2, focal atrophy pattern was found: 164 out of 1829 PSCs showed a significant difference. Hippocampus regions were highly involved in this subtype. Further evidence of the anatomical heterogeneity exhibited by the two subtypes of AD can be seen when the PSCs that MAGIC utilizes in its classification boundary were projected onto two dimensions using multidimensional scaling. Subtype 1 and 2 in blue and green, respectively exhibit unique divergences away from CN along with two directions, predominately described by the presence and absence of cortical atrophy patterns (Fig.~\ref{fig3}C).      
\subsubsection{Individualized Classification via the Two-face Polytope} The nature of MAGIC allows not only for clustering but also for classification via the convex polytope. For a fair comparison, we randomly split AD patient in the training set into 2 splits with the same ratio as the two subtypes found in MAGIC (134/57). Taking the PSCs as features (\emph{K} = 35) when MAGIC converged, two linear SVMs were independently run for CN vs first split and CN vs second split of AD to construct a polytope as in MAGIC. Fig.~\ref{fig3}D showed that MAGIC (\(0.82\pm0.03\)) and the permutation linear SVMs (\(0.80\pm0.03\)) obtained comparable results. Of note, since no nested CV for hyperparameter searching or feature selection was performed, the accuracy here is lower compared to state-of-the-art \cite{AD-ML,AD-ML-DTI}.   
\section{Conclusion}
In the current study, we proposed a novel method, MAGIC, for parsing disease heterogeneity and demonstrated its superiority over HYDRA. The application to AD found two robust clinically different subtypes, thus highlighting the potential of MAGIC in the analysis of the heterogeneity of brain diseases.

\end{document}